%% file: acl_latex.tex
\documentclass[11pt]{article}

\usepackage[final]{acl}

\usepackage{times}
\usepackage{latexsym}

\usepackage[T1]{fontenc}

\usepackage[utf8]{inputenc}

\usepackage{microtype}

\usepackage{inconsolata}

\usepackage{graphicx}

\newcommand{\huggingface}{\raisebox{-1.5pt}{\includegraphics[height=1.05em]{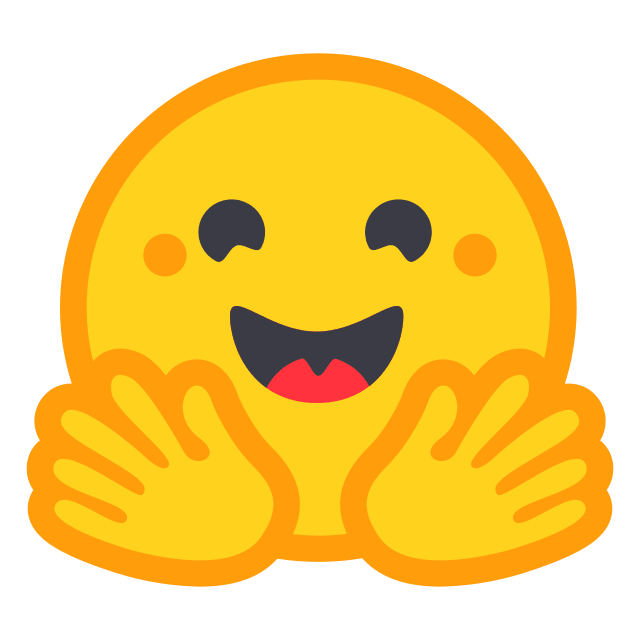}}}

\title{Model in Distress: Sentiment Analysis on  French Synthetic Social Media}

\author{Pierre-Carl Langlais$^1$ ~~~ Pavel Chizhov$^{1,3}$ ~~~ Yannick Detrois$^{1,4}$ ~~~ Carlos Rosas--Hinostrosa$^1$ \vspace{1mm} \\
\textbf{Ivan P. Yamshchikov}$^{1,3}$ ~~~~ \textbf{Bastien Perroy}$^2$\vspace{2mm} \\
$^1$PleIAs, Paris, France ~~~~ $^2$Passenger Cognition Lab, RATP Group, Paris, France \\
$^3$CAIRO, THWS, Würzburg, Germany ~~~~ $^4$EPFL, Lausanne, Switzerland \\
\small{
    \textbf{Correspondence:} \href{mailto:pierre-carl@pleias.fr}{pierre-carl@pleias.fr}
  }
  }

\usepackage{verbatim}
\usepackage{booktabs}

\definecolor{new_purple}{RGB}{153, 51, 255}
\definecolor{new_magenta}{RGB}{255, 51, 153}

\begin{document}
\maketitle

\input{tex/00_abstract}

\input{tex/01_introduction}

\input{tex/02_related_work}

\input{tex/03_methods}

\input{tex/04_results}

\input{tex/05_discussion}

\input{tex/06_conclusion}


\bibliography{custom}

\input{tex/XX_appendix}
\end{document}

%% file: tex/00_abstract.tex
\begin{abstract}

Automated analysis of customer feedback on social media is hindered by three challenges: the high cost of annotated training data, the scarcity of evaluation sets, especially in multilingual settings, and privacy concerns that prevent data sharing and reproducibility. We address these issues by developing a generalizable synthetic data generation pipeline applied to a case study on customer distress detection in French public transportation. Our approach utilizes backtranslation with fine-tuned models to generate 1.7 million synthetic tweets from a small seed corpus, complemented by synthetic reasoning traces. We train 600M-parameter reasoners with English and French reasoning that achieve 77-79\% accuracy on human-annotated evaluation data, matching or exceeding SOTA proprietary LLMs and specialized encoders. Beyond reducing annotation costs, our pipeline preserves privacy by eliminating the exposure of sensitive user data. Our methodology can be adopted for other use cases and languages.
\begin{center}
\small\huggingface~\href{https://huggingface.co/datasets/DistressedModel/Distressed-Model-Data}{\path{DistressedModel/Distressed-Model-Data}}
\end{center}
\end{abstract}

%% file: tex/01_introduction.tex
\section{Introduction}

Large companies and services handle a substantial volume of reviews and social media mentions. Social media has become a primary means of filing complaints due to the need for unfiltered, independent feedback and ease of use~\cite{SocialMediaToday2014,Causon2023,article}. This convenience has driven a corresponding increase in the volume of complaints. For example, by 2014, complaints in the United Kingdom had doubled compared to 2013, with 31\% expressed through social media~\cite{ISTANBULLUOGLU201772}. Social media also provides anonymity and eliminates direct confrontation, enabling customers to express frustration while exerting perceived public pressure that private reviews may lack~\cite{https://doi.org/10.1111/1467-8551.12928}. Furthermore, if customers complain offline or in private, they may then opt for a second round of complaint online~\cite{10.1108/IJRDM-03-2020-0089}.

As customer expectations around social media engagement grow, response time has become critically important~\cite{Sigurdsson2021}. For instance, it was recently shown to influence the chances of repurchase~\cite{ISTANBULLUOGLU202411}. Therefore, it is crucial for a company to quickly detect the issues and react accordingly by implementing a service recovery~\cite{https://doi.org/10.1111/1911-3838.12339}. When services operate at scale, and social media interactions exceed manual processing capacity, automated tools such as language models offer a practical solution.

Language models of different sizes and with different design purposes are being developed daily. Proprietary large language models (LLMs) might act as agents for such purposes, while an alternative solution would be specialized small language models. The benefits of LLMs typically include their versatility and ability to understand complex contexts, while specialized small language models are favored for their robustness and, what is especially relevant to public institutions, for the traceability of open-source solutions and an opportunity for autonomous deployment.

A crucial part of automatic review evaluation is sentiment analysis, which typically involves detecting a wide range of emotions and distress to identify the purpose and sentiment of the user messages and classify them into categories required by the business. The models for such a classification task require training on a substantial portion of representative data and reliable annotations, which are often costly to obtain. Furthermore, the trained models need to be properly evaluated, which is often problematic due to the specificity of each case and the lack of comprehensive benchmarks, especially for non-English languages. Even though modern machine translation systems are very advanced, simply translating all the training data into the required language would degrade data quality~\cite{Arnett2025}, especially given the domain's complex style.

Synthetic data generation has recently emerged as a promising method for training specialized models in regulated sectors with the lack of annotated data~\cite{hughes_improving_2024, mendes_synthetic_2025, meyer_nemotron-personas_2025, tan_synthesizing_2025}. Pipelines provide privacy and compliance by design, as personal or sensitive data is never exposed; instead, it is substituted with realistic simulations.

In this work, we introduce one of the first large-scale synthetic pipelines for customer reviews on social media. The pipeline can be applied across various domains and languages to address the aforementioned challenges. We demonstrate its effectiveness on the example of Paris public transport. Our key contributions include:
\begin{itemize}
    \item We collect a small dataset of tweets annotated by both native speakers and LLMs and aggregate these annotations into reliable labels.
    \item We present a methodology for creating a synthetic pipeline for review generation using backtranslation and synthetic reasoning, which is applicable to other domains, and release a large collection of synthetic tweets\footnote{We release the anonymized synthetic tweets on HuggingFace: \huggingface~\href{https://huggingface.co/datasets/DistressedModel/Distressed-Model-Data}{\texttt{DistressedModel/Distressed-Model-Data}}.}.
    \item We train small reasoning models using synthetic data and evaluate them using human-annotated data, reaching a performance comparable to or better than that of modern LLMs.
\end{itemize}

%% file: tex/02_related_work.tex
\section{Related Work}

Previous sentiment analysis research was largely concentrated on English data. TweetEval~\cite{barbieri2020tweeteval} combined several SemEval tasks into a compound benchmark for sentiment analysis~\cite{rosenthal2017semeval}, emotion recognition~\cite{mohammad2018semeval}, stance~\cite{mohammad2016semeval}, irony~\cite{van2018semeval}, hate speech~\cite{basile-etal-2019-semeval}, and offensive language detection~\cite{zampieri2019semeval}, as well as emoji prediction~\cite{barbieri2018semeval}. All the subsets were exclusively in English. Despite the initial prevalence of small encoder models applied to this task, it was recently used to evaluate large language models~\cite{AIMultiple2025}.

Previous efforts for non-English data collection and classification included tweet sentiment analysis in 14 African languages~\cite{muhammad-etal-2023-afrisenti}, multi-modal analysis in 21 languages~\cite{thakkar-etal-2024-m2sa}, and other approaches in Eastern European languages~\cite{filip_2025_17723755}, Bengali~\cite{sazzed-2020-cross}, Indonesian~\cite{geni2023sentiment}, and Hausa~\cite{ALIYU2025100330}.

\paragraph{Encoder-style solutions.} Typical solutions for the classification tasks of such sort are small BERT-like models~\cite{HOQUE2024100091,MOURA2024103688}, or classical machine learning approaches to building embeddings (TF-IDF, Bag of Words, Word2Vec, etc.) and classification algorithms (Logistic Regression, Random Forest, etc.)~\cite{AGRAWAL2025103407}, as well as custom pipelines combining both approaches~\cite{Mudarakola2025}.

\paragraph{Decoder-style solutions.} There has also been a growing line of work where decoder-style language models are used for sentiment analysis and review classification~\cite{suen2025small,elmitwalli2024enhancing}. Even though these approaches are more computationally intensive, which is partially mitigated by recent advancements in small language model development~\cite{gemmateam2025gemma3technicalreport,yang2025qwen3technicalreport}, they often provide additional features such as custom prompts~\cite{suen2025small} and reasoning~\cite{AHMED2026103668}.

%% file: tex/03_methods.tex
\section{Task and Motivation}
\label{sec:motivation}

\begin{figure*}
    \centering
    \includegraphics[width=\linewidth]{figures/ratp-new.png}
    \caption{Synthetic pipeline and distress detection model training. \textbf{Left:} auxiliary model fine-tuning. \textbf{Right:} synthetic pipeline for generating synthetic tweets and training the distress detection model. All LLM-generated data are shown in \textcolor{new_purple}{purple}. All pipeline-generated synthetic data are shown in \textcolor{new_magenta}{pink}.}
    \label{fig:pipeline}
\end{figure*}

In this work, we analyze public transport reviews on Twitter in French and aim to build a \textbf{distress detection} tool that enables support agents to quickly monitor issues expressed on social media. While the task is similar to an ordinary classification pipeline, we approach it in a non-standard way by building a synthetic tweet generation pipeline and training small reasoning models. This is motivated by the following factors:
\begin{itemize}
    \item \textbf{Imbalance:} operational classification of distress is different from the abstract definition and concerns a very small selection of tweets. Using only organic data would put a ceiling on the capabilities of small models. 
    \item \textbf{Data drift:} even though a large historical dataset of 30 million tweets collected from 2012 onward (see Appendix~\ref{app:data}) could be leveraged to mitigate data scarcity, there have been substantial changes in user expression and collection design. Thus, any models trained on these data might perform suboptimally on more recent data.
    \item \textbf{Data sensitivity:} tweets have always been a gray area regarding GDPR, as, despite being public expressions, they might contain identifiable or sensitive information on private persons. Moreover, monitoring also typically includes strictly private user data (application feedback, phone calls) that could never be disclosed for training.
    \item \textbf{Generalization:} The monitoring infrastructure anticipates a range of extreme events that never occurred in the historical data, which can be handled by language models with generalist knowledge and reasoning capabilities.
\end{itemize}

\noindent Given the unique challenge of realistic, primarily non-English tweet generation, we build our synthetic pipeline around custom finetuned models. Thus, we approach this problem via \textbf{specialist distillation}~\cite{deepseek-v32}, when a specialized model is trained to produce realistic synthetic data, which is then used to train the final model.

\section{Methodology}
\paragraph{Data.}
The original dataset comprises more than 30 million tweets related to French public transport between 2012 and 2025, from over 5 million users. Data acquisition during this period proceeded in separate phases. During the \textbf{first phase} (2012 to June 2023), a Twitter research license permitted large-scale data scraping of tweets that mention French public transport or are replies to threads referencing it. This ensures exhaustive coverage but also produces noisy data, including possibly unrelated tweets. This phase accounts for the largest part of our data, with peak monthly tweet counts exceeding 1.5 million. The \textbf{second phase} (November 2024 and onward) followed Twitter's acquisition and evolution of social media infrastructure. A professional API license was acquired to collect the most recent and relevant tweets in smaller volumes (on average 30\,000 tweets per month). While the majority of tweets are in French (66.3\%), English is also represented (15.3\%), with other less prominent languages making up the remaining fraction. The complete details about the dataset are presented in Appendix~\ref{app:data}.

\input{tex/results_table}

\paragraph{Synthetic pipeline.}

The core of our synthetic pipeline (Figure~\ref{fig:pipeline}) relies on \textbf{backtranslation}. We use Gemini 3 Pro~\cite{Doshi2025Gemini3ProVision} as a convenience to generate our  3\,000 seed annotations (see prompts in Appendix~\ref{app:prompts}) using a sample from the second phase of our dataset. We use a proprietary model for this generation, although recent capable open-weight models can serve as an effective substitute~\citep{sutro2026synthetic,epoch2025openweights}. At the initial stage of the pipeline (left part of Figure~\ref{fig:pipeline}), we annotate a set of tweets with distress labels and a set of features, combining linguistic characteristics, emotional valence, and general contextual information (event, location, and any identification of the author). We also generate synthetic reasoning traces that lead to tweets from their annotations, and vice versa. We then use these data to train specialized annotation and reasoning models that will be used as parts of the synthetic pipeline. For reasoning traces, we use both English and French in separate pipelines, yielding two different reasoner models. To stimulate diversity for the linguistic annotation model, we also add a portion of tweets unrelated to distress. We also train the tweet generation model on a backtranslation-style combination of annotations, reasoning traces, and original tweets. All four auxiliary models are fine-tuned versions of Gemma 3 12B Instruct~\cite{gemmateam2025gemma3technicalreport}.

After this, we use a filtered collection of first-phase tweets as seed to generate a dataset of 1\,737\,797 synthetic tweets using the fine-tuned models (right part of Figure~\ref{fig:pipeline}). We separately validate the variety of generated tweets due to the risk that the generation model may reproduce the seed tweets. Of the generated tweets, 14\,429 (0.83\%) were close to the original tweets, as evaluated with \texttt{SequenceMatcher} from \texttt{difflib} using a similarity threshold of 65\%. We filter these entries out before the training of the final model. We combine the resulting 1\,723\,368 synthetic tweets with generated reasoning traces and SYNTH\footnote{\url{https://huggingface.co/datasets/PleIAs/SYNTH}}, an open-source pre-training dataset for generalist reasoning, into a training dataset for our final model, a decoder model with Llama-style architecture and 600 million parameters (see architecture details in Appendix~\ref{app:model}). The training dataset results in 1\,010\,748\,472 tokens, over which we train for three epochs on four NVIDIA H100 GPUs with an effective batch size of 262\,144 tokens: context length of 2\,048, per-GPU-batches of size 8, and 8 gradient accumulation steps.

\begin{figure}
    \centering
    \includegraphics[width=\linewidth]{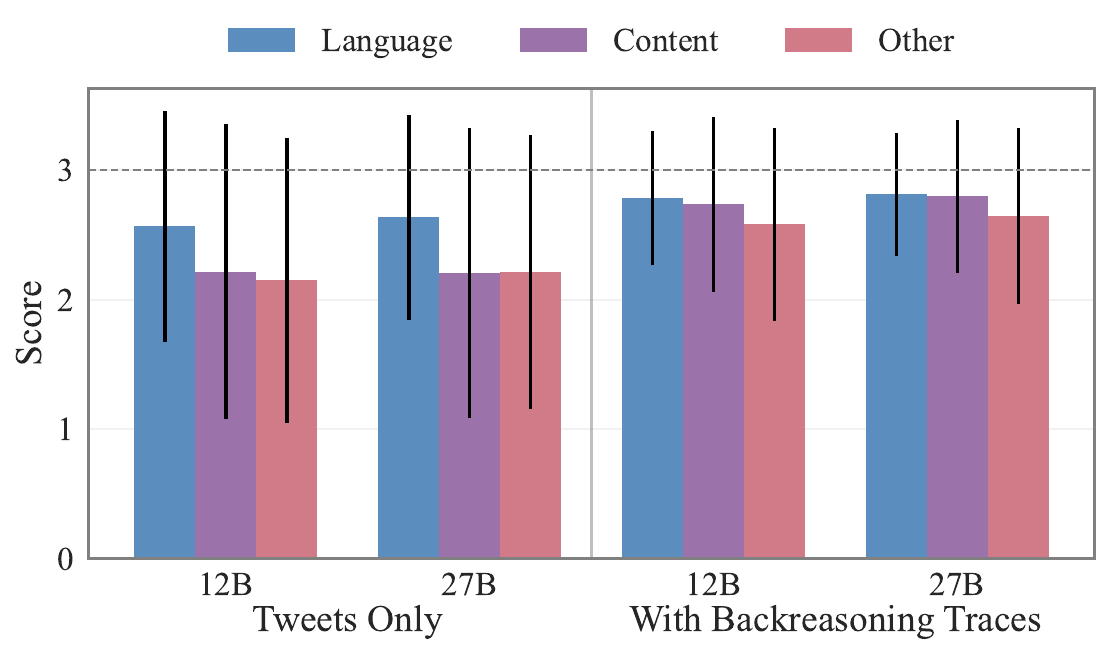}
    \caption{Evaluating Gemma models fine-tuned for tweet generation. Bars represent standard deviation.}
    \label{fig:eval_gemma}
\end{figure}

\paragraph{Evaluation sample.}

From the second-phase subsample of the dataset, we select 2\,000 relevant tweets using annotations by three LLMs. The selected tweets do not intersect with the fine-tuning sample in the synthetic pipeline. We then slice the selected sample into 40 batches of 50 tweets and hand them to a group of annotators, so that each tweet receives from 9 to 11 binary distress annotations, resulting in an average of 10.2 annotators per tweet (see Appendix~\ref{app:annotations} for the complete annotation details). We aggregate the labels through majority voting, resolving ties in favor of the distress label. We leverage this sample for the evaluation of the final distress detection model.

%% file: tex/results_table.tex
\begin{table*}[tbp]
\small
\centering
\begin{tabular}{lcccccc}
\toprule
\textbf{Model} & \textbf{Parameters} & \textbf{Fine-tuning} & \textbf{Accuracy } & \textbf{Precision } & \textbf{Recall } & \textbf{F1 } \\
\midrule
Claude Sonnet 4.5 & --- & --- & 72.2 & 75.8 & 62.8 & 68.7 \\
Gemini 2.5 Pro & --- & --- & 71.5 & 71.1 & 69.5 & 70.3 \\
Mistral 3 Large & 675B & --- & 78.5 & 88.6 & 64.1 & 74.4 \\
Qwen3.5 & 397B & --- & 67.5 & \textbf{91.3} & 36.8 & 52.5 \\
\cmidrule(ll){2-7}
Distress Reasoner (ours) & 600M & Synth. tweets, EN reasoning & 78.0 & 80.7 & 71.9 & 76.0 \\
Distress Reasoner (ours) & 600M & Synth. tweets, FR reasoning & \textbf{79.5} & 82.2 & \textbf{73.7} & \textbf{77.7} \\
\midrule
ModernBERT Large & 395M & Synth. tweets & 79.1 & 81.0 & \textbf{74.4} & 77.6 \\
CamemBERTv2 Base  & 110M & Synth. tweets & \textbf{79.3} & \textbf{81.6} & 74.3 & \textbf{77.7} \\
\bottomrule
\end{tabular}
\caption{Accuracy (\%), Precision (\%), Recall (\%), and F1 evaluations for distress detection. We compare LLMs (top rows), our decoder models (middle rows), and fine-tuned encoders (bottom rows). Our decoder models are trained separately with English (EN) and French (FR) reasoning. The best scores among decoders and encoders are \textbf{in bold}.}
\label{tab:evaluation_results}
\end{table*}

%% file: tex/04_results.tex
\section{Experimental Results}

\paragraph{Auxiliary model ablation.} We validate the tweet annotation model using Gemini 3 Pro as a judge to assess how a generated tweet aligns with the generation instructions (see evaluation instructions in Appendix~\ref{app:judge}). We present a comparison of two Gemma versions (12B and 27B), fine-tuned with and without synthetic reasoning traces, in Figure~\ref{fig:eval_gemma}. The results show that synthetic reasoning traces improve the correspondence to factual labels in the generation instructions, while language quality remains high in both modes. The difference between the 12B and 27B models is negligible; therefore, we utilize the 12B version for speed.

\paragraph{Distress detection.} We present the results of evaluating on the human-annotated sample of tweets in Table~\ref{tab:evaluation_results}. We compare our small models to proprietary LLMs: Claude Sonnet 4.5~\citep{anthropic2025sonnet45} and Gemini 2.5 Pro~\citep{comanici2025gemini25pushingfrontier}, and large open-weight LLMs: Mistral 3 Large~\citep{mistral2025mistral3} and Qwen3.5 397B~\citep{qwen2026qwen35}. We evaluate our models with low temperature and repetition penalty values (both set to $0.1$) and treat parsing failures as negative predictions. Despite their small size, the newly trained specialized models achieve competitive performance on the constructed task, comparable to or better than heavy multilingual LLMs with reasoning capabilities. Mistral 3 Large achieves the closest performance, possibly due to better adaptation to real-world French data, yet has inferior recall, which is crucial in complaint-handling scenarios. Qwen3.5 has the highest precision, but the lowest recall. We hypothesize that, although the model has multilingual capabilities, it is not well-adapted to complicated non-English Twitter slang.

We separately fine-tune two encoder models on our synthetic training set: ModernBERT Large~\citep{warner-etal-2025-smarter} and CamemBERTv2 Base~\citep{antoun2024camembert20smarterfrench}\footnote{We release the fine-tuned encoder-style models as \huggingface~\href{https://huggingface.co/DistressedModel/modernbert-distress-signal}{\texttt{DistressedModel/modernbert-distress-signal}} ~~and \huggingface~\href{https://huggingface.co/DistressedModel/camembert-distress-signal}{\texttt{DistressedModel/camembert-distress-signal}}}. Both models performed comparably to our best decoder model, with CamemBERT performing slightly better, likely due to the better adaptation to French content. This shows that our synthetic pipeline is also suitable for training specialized encoders; however, encoder models lack reasoning traces, which are highly practical for real-world scenarios. We further discuss this in Section~\ref{sec:discussion}.

%% file: tex/05_discussion.tex
\section{Discussion}
\label{sec:discussion}

Our results demonstrate that a small set of relevant real-world data can be used as a seed for the synthetic generation of reliable data, including synthetic reasoning. This approach made it possible to share the training data, ensuring better standards of reproducibility and fostering similar work on sensitive data resources or annotation. The data generated at scale exhibits diversity and can then serve as a training set for a classifier that will be suitable for training a real-world distress detection model that achieves state-of-the-art performance comparable to or better than that of proprietary LLMs. The demonstrated methods can be used to scale up the training data in conditions of low availability of annotated real-world data. The resulting model can then be autonomously deployed without reliance on a proprietary commercial API and efficiently process large portions of reviews due to its small size (600 million parameters). 

The comparable performance of encoder-based models demonstrates that the pipeline can be used for encoder-based scenarios as well. However, a decoder-based reasoner offers two advantages over a scalar encoder. First, conditioning the label on an explicit reasoning trace improves correspondence with the fine-grained content annotations requested at generation (Figure~\ref{fig:eval_gemma}). Second, the reasoning traces persist during model inference as usable outputs for agents acting on each decision. This aligns with the argumentative view of reasoning, on which reasoning is fundamentally an artifact produced to be evaluated by others~\citep{Mercier_Sperber_2011}: an opaque score offers nothing to scrutinize, whereas a trace exposes the inferences a downstream agent can accept, contest, or override. In a cognitive task analysis of three rail control rooms, \citet{dadashi_modelling} make the parallel operational point that automation should support exploration of the reasoning behind its decisions rather than expose only an alert. The same trace also externalizes which patterns the model has learned to associate with distress, complementing the privacy-preserving design of Section~\ref{sec:motivation} with the form of transparency that European frameworks such as the GDPR and the AI Act increasingly require of automated decision systems acting on personal data. In the public transport setting, the actions taken on the basis of a flagged tweet can range from a routine acknowledgment to consequential interventions such as dispatching emergency staff or evacuating passengers from a train stalled in a tunnel; in this regime, the trace lets a support agent inspect the cited spans rather than accept or reject an opaque score. This is an affordance no BERT-style classifier provides.

\section{Conclusion}

Our work provides a proof of concept that, under a specialist synthetic pipeline, a small language model with reasoning can achieve performance comparable to or better than state-of-the-art large models, both proprietary and open-weight. Such a model is beneficial not only from the perspective of specialization and task-tailoring, but also enables fast processing of large volumes of requests and fully autonomous deployment. The decoder design with reasoning traces also offers practical benefits over encoder-based models. The demonstrated pipeline is relatively universal and can be adapted to various domains and languages, given a small set of seed data and a capable auxiliary language model that can generate the seed synthetic data in the required language.

%% file: tex/06_conclusion.tex
\section*{Limitations}

Our work demonstrates results only on one classification characteristic (distress) and only in a specific domain and language (public transport reviews in French). Thus, the model's application to other tasks and domains would require separate fine-tuning and additional synthetic data generation. 

The demonstrated approach requires a certain amount of real-world data as a seed for fine-tuning auxiliary models and generating synthetic data. Thus, the pipeline is not suitable for cases where real-world data is unavailable. Furthermore, since our original annotations and synthetic reasoning were generated by Gemini, this approach may not be suitable for private real-world data. In such cases, practitioners would need a locally deployed LLM for auxiliary data generation; however, the results from such a setup might be just as good given the right model.

We used Gemma as a base model for synthetic fine-tuning, which is essentially strong in terms of multilinguality. However, if a new pipeline requires expertise in an extremely low-resource language, a suitable fine-tuning model should be selected. This is especially important when, as in our case, the task requires strong generalizability to the unseen cases and understanding of complicated language nuances and slang.

\section*{Acknowledgements}
The computational resources for this work were provided by Jean Zay supercomputer (project EuroSynthPlay, compute grant AD011016886R1).

%% file: tex/XX_appendix.tex
\appendix

\section{Data}
\label{app:data}

\begin{figure*}[!t]
    \centering
    \includegraphics[width=0.95\textwidth]{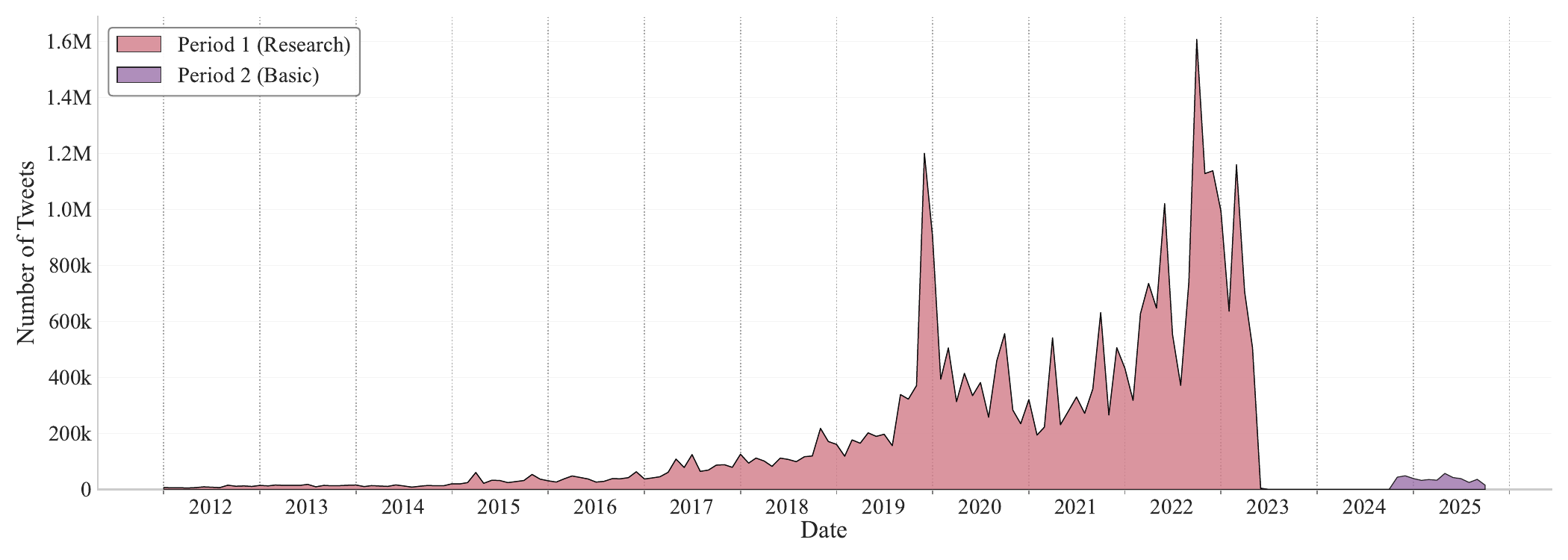}
    \caption{Monthly tweet volume by collection period between 2012 and 2025. }
    \label{fig:monthly_tweets}
\end{figure*}

\begin{figure*}[!t]
    \centering
    \includegraphics[width=0.95\textwidth]{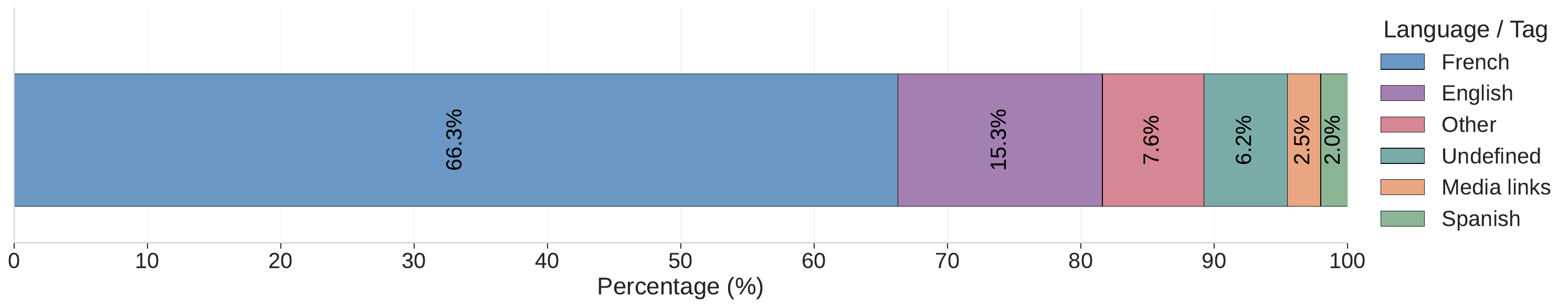}
    \caption{Tweet language distribution in the historical period (Period 1). }
    \label{fig:language_stats}
\end{figure*}

In the first collection period between 2012 and 2023, the volume of tweets increased steadily over the years despite an invariable collection scheme, reflecting the growing importance of transport on social media (see Figure~\ref{fig:monthly_tweets}). With the 2023 API changes, the research license was switched to a basic license with more restrictive collection criteria. Although the new scheme produces a smaller dataset (about 30\,000 tweets compared with nearly one million by the end of Period 1), the remaining volume is still sufficient to capture the most important information and identify distress signals. Data from the period between the two phases was not used in this work. 

Approximately two-thirds of the dataset consists of French tweets, with the remainder comprising multiple languages, which supports multilingual capabilities (Figure \ref{fig:language_stats}). The second-largest language is English, followed by Spanish, with other languages grouped under \textit{Other}, each representing less than 2\% of the data. Certain tweets that contain only, or primarily, emojis, media, mentions, hashtags, or links have an undefined language attribute (6.2\%). Since 2022, new tags have been introduced to more precisely categorize undefined language attributes. For example, the \textit{Media Links} tag for tweets containing only media links, which account for an additional 2.5\% of the data.

\section{Tweet Annotations}
\label{app:annotations}
We curated a challenging annotation dataset of real 2\,000 tweets. Given the rarity of distress signals, we used an initial selection using the annotations from three LLMs. We include the full evaluation set with dehydrated Twitter IDs\footnote{\huggingface~\href{https://huggingface.co/datasets/DistressedModel/Distressed-Model-Data}{\path{DistressedModel/Distressed-Model-Data}}}.

We construct our ground truth evaluation dataset using a crowdsourcing strategy, recruiting a cohort of annotators to label the data. The objective is to determine whether a message expressed distress and to quantify the extent of experiential cues provided by the passenger. A total of 2\,000 tweets are distributed into batches. Each participant is assigned a batch of 50 tweets and is required to provide a binary classification of distress (Yes/No) and an assessment on a Likert scale of the degree of experiential detail. To mitigate order effects, participants first complete the distress classification for their assigned batch in a randomized order, followed by the experiential cues assessment for the same batch in a newly randomized order.

To ensure data quality, we implement two standard procedures to detect Insufficient Effort Responding (IER). First, following the recommendations of \citet{meade2012identifying} for multiple attention check designs, we embed four attention check items (e.g., ``Please select `Yes' for this item'') within the survey. Participants who fail more than one check (i.e., fewer than 3 correct answers) are excluded; this criterion led to the exclusion of 4 participants from the initial 415. Second, we screen for rapid responding using the method described by \citet{huang2012detecting}. We calculate the median response time for each participant across all 100 item-level variables. Respondents with a median response time below 2 seconds are flagged as ``speeders,'' as this duration is considered the cognitive minimum required to process a survey item meaningfully. This median-based approach is robust to occasional outliers while effectively identifying consistent careless patterns. This threshold eliminated four additional participants (1.4\%).

The final analytical sample consists of 407 participants recruited via convenience sampling. We employ a consensus strategy, using panels of nine to eleven distinct evaluators per message, specifically aiming for a diverse cross-section of usage intensity. The cohort is split between regular riders ($n=196$, 48.2\%) and occasional riders ($n=211$, 51.8\%). We define regular riders as those traveling daily or almost daily ($n=88$), several times a week ($n=63$), or several times a month ($n=45$). Conversely, the occasional group comprises those who travel several times a year ($n=121$) or less frequently ($n=90$). This heterogeneity is strategic. While annotators may lack formal expertise in text analysis, their personal history of navigating the network grants them a critical ``situated expertise.'' This personal experience with the transport environment enables them to intuit the complex, often implicit experiential cues.

Overall agreement among annotators is 80.50\% (± 16.70\%), which is within expectations given the focus of evaluation on complex cases.

\section{Architecture and Training}
\label{app:model}

In Table~\ref{tab:model-training-config}, we present the main model architecture and training hyperparameters.

\begin{table}[!t]
\centering
\small
\begin{tabular}{ll}
\toprule
\textbf{Category} & \textbf{Setting} \\
\midrule
\multicolumn{2}{c}{{Model Architecture}} \\
\midrule
Number of layers           & 80 \\
Hidden size                & 800 \\
Intermediate (FFN) size    & 2048 \\
Attention heads            & 10 \\
Key--value heads           & 5 \\
Activation function        & SiLU \\
Max sequence length        & 2048 \\
Positional encoding        & RoPE ($\theta = 10{,}000$) \\
Normalization              & RMSNorm ($\epsilon = 1\times10^{-5}$) \\
Embedding tying            & Enabled \\
Vocabulary size            & 65{,}536 \\
Precision                  & \texttt{bfloat16} \\
\midrule
\multicolumn{2}{c}{{Training Setup}} \\
\midrule
Optimizer                  & AdamW \\
Adam $\beta_1, \beta_2$    & 0.9, 0.95 \\
Weight decay               & 0.01 \\
Gradient clipping          & 0.5 \\
Learning rate              & $5\times10^{-4}$ \\
Warmup style             & Linear \\
Decay style                & Linear \\
Warmup steps               & 500 \\
Decay start step           & 3000 \\
Minimum learning rate      & $6\times10^{-6}$ \\
Training steps             & 4000 \\
Micro-batch size           & 4 \\
Gradient accumulation      & 8 \\
Effective sequence length  & 2048 \\
\midrule
\multicolumn{2}{c}{Parallelism} \\
\midrule
Data parallelism (DP)      & 4 \\
Tensor parallelism (TP)    & 1 \\
Pipeline parallelism (PP)  & 1 \\
ZeRO stage                 & 0 \\
\bottomrule
\end{tabular}
\caption{Model architecture and training configuration.}
\label{tab:model-training-config}
\end{table}

\section{LLM-as-a-Judge}
\label{app:judge}

For the evaluation of how generated tweets correspond to the generation instructions, we use the following rubrics for Gemini 3 Pro~\citep{Doshi2025Gemini3ProVision}, which is used as a judge:

\begin{itemize}
    \item \textbf{Language}: overall linguistic quality, register, code-switching if applicable, and formality. 
    \item \textbf{Content}: topic and thematic accuracy, the described issue type, the mentioned audience or target. 
    \item \textbf{Other attributes}: emotions, tone, irony, urgency, stance, intent, intensity, and any other stylistic or pragmatic features.
\end{itemize}

Each criterion is scored on a scale from 0 to 3 with labels:  
\begin{itemize}
    \item 3 = good compliance (expected correct behavior)
    \item 2 = minor deviations
    \item 1 = major deviations
    \item 0 = complete failure or irrelevance
\end{itemize}

\section{Generation Prompts}
\label{app:prompts}

We present prompts used for the generation of the fine-tuning data with an LLM (Gemini 3 Pro;~\citealp[]{Doshi2025Gemini3ProVision}) in Figures~\ref{fig:prompt-1},~\ref{fig:prompt-2}, and~\ref{fig:prompt-3}.

\section{LLM Usage Statement}

In this work, we used AI tools for paraphrasing and grammar correction (Grammarly) and for data visualization design (Claude).

\input{tex/prompts/reasoning-generation}

\input{tex/prompts/linguistic-features}

\input{tex/prompts/draft-generation}

\input{tex/prompts/reasoning_example}

%% file: tex/prompts/reasoning-generation.tex
\onecolumn

\begin{figure}[t]
\scriptsize
\centering
\begin{verbatim}
You are a **clinician-like annotator** assessing a single tweet for potential **distress and psychometric states**.
You work **bilingually (French/English)**: interpret and quote French text accurately, but write all analysis in English.
Use concise, evidence-based prose.

You have been submitted the following tweet:
{tweet}

You also no in advance the final judgement:
{judgement}

Now try to write a reasoning workflow according to the guidelines below that would converge to this final judgement.

### 1) Pragmatics
* Contextualize: is it a 
* Identify tone and stance: literal report / complaint / rhetorical slogan / meme or sarcasm / hyperbole / bureaucratic or formal.
* Who speaks about whom? (self vs. other vs. unspecified).
* Temporal index: now / recent / past / unknown.
* Quote 1–3 exact spans that anchor these judgments.

### 2) Distress

**Goal:** Determine whether any form of distress is present.

* Decide: **present / absent / external**

**Criteria for present:**
* Author is a victim or a direct witness to physical violence or danger. This could be caused by other travellers or agents. 
Common issues include being thrown on the rails, attacked by a third party, etc.
* Author is a victim or a direct witness of acts sexual violence, not only assault but also harassment, exhbitionnism.
* Author is a victim or a direct witness health issues, including faintness ("malaise"), breathing (especially is ashtma), 
overheating, body compression in highly crowded space.
* Graphic description of violence likely witnessed or even endured by the author (blood, members sectioned, etc.)
* Strong potential for violence or whole situation degenerating. Typically someone having a very disturbing behavior (under drugs, displaying 
mental issues, claiming some terrorist intent, being armed, laying something looking like a bomb)
* Immediate and unusual infrastructure risks, even if nothing happened yet: unsecurized rails, flooding, ceiling falling, invasive species 
("punaises), etc.
* Collective distress situation including heat, crowding, discomfort. This may spill into potential or actual fainting, especially involving 
fragile people (old, etc.) or collective panic/psychosis.
* Author being a direct witness to train evacuation, (having to break windows, walk on tunnels) or train getting stuck for a long time 
(up to hours) with indirect health risks.
* Author showing signs of breakdown, even if the cause might not be dramatic in itself: crying/screaming, disorientation. This can be 
expanded by somatic, cognitive, or fragmentation cues.
* Generic yet genuine cry for help, even if details are unclear. So long as the author seems serious and currently enduring this situation

**Criteria for absent:**
* Political rhetoric or commentary with no self-referential distress.
* Light inconvenience expressed through meme, hyperbole, or sarcasm.
* References to self-harm or even suicide ideation that are clearly emphatic.
* Service disruption without spillover risk. This include being stuck in a tunnel for a shortwhile, losing a connection, etc.
* Generic complaint about degraded environment ("racailles").
* Potentially accute annoyance (noise, smells) but without traumatic consequences
* Losing personal objects (bag, wallet, computer) in the subway, unless this translate to accute personal trauma

**Criteria for external:**
* Reference to a clear distress situation but **¨reported** through external sources.
* Downstream consequence of an actual distress situation ("malaise voyageur" causing delays but not happening nearby)
* Any past and resolved event (especially if the person was saved in the end)
* Publication by a media/journalist or other third party.
* Video coverage

Provide a **one-sentence justification** including **verbatim span(s)**.

### 3) Semiotic potential
As a last test, you have to evaluate whether the tweet alone contain enough semiotic signals for a subsequent psychometric assessment.

The cues can be of different nature:
*Explicit relation of a traumatic event.
*Description of inner or supposed mental state.
*Description of psychosomatic symptoms.
*Traumatic connotations in the text itself. This obviously include high emotional valence but even signs of dissociation (like the person
was agressed and write in a super cold tone)

Here your final judgment is a grade, from 1-10.
* 1–3 = figurative or faint
* 4–6 = partial or indirect cues
* 7–10 = explicit, concrete, well-anchored evidence

Produce all results in this order. Never mention that you know the end result in advance, just try to converge to it.

### Text stance & pragmatics ###
[Initial analysis/exploration]

### Distress ###
[Is this relevant for distress? Why? Quote spans. Final answer either **present**, **absent** or **external**]

### Semiotic potential ###
[Primary class + micro-justification with quotes.]
\end{verbatim}
\caption{Prompt for distress detection and reasoning.}
\label{fig:prompt-1}
\end{figure}

%% file: tex/prompts/linguistic-features.tex
\begin{figure}[t]
\scriptsize
\centering
\begin{verbatim}
I would like to annotate the following tweet:
{tweet}

The annotation is strictly focused on the language analysis. It used the following features.

<event>
A description of the event related in the tweet in English. Should be very brief and hold in one sentence.

<location>
If indicated, name of the location(s) where the event take place. Could be typically subway stations, bus stops, streets, etc.

<time>
If indicated/specified, time of the day or the year the event take place. You only need to be as precise as is mentioned in the text: 
could be just hour (9 am) or a complete timestamp.

<persona>
If specified, a short description of what the author of the tweet could be based on text clues. Typically, a middle-aged business person 
from Korea, a young French muslim woman, etc. Leave blank if you don't know anything more than a frequent commuter in Paris.

<language_register>
Standardized values can be either:
formal: official complaints, institutional tone, requests for attestations
standard: standardized conversational tone in French, neutral tone, proper grammar,
casual: colloquial form of French for online discussions ("mdr", "vzy", contractions like "j'suis", "y'a")
street: argot, verlan like "ouf", "chelou"
dialectical: regional variety of French, not necessarily in France (also include pidgin etc.)

<language_quality>
standard: generally correct French
imperfect: a few typo, still acceptable for conversational style
poor: hard to parse, showing visible struggle with written French

<language_switching>
If valid, simply list the languages that used in lowercase. If French and English: french, english etc.

<emotion>
The primary emotion conveyed by the tweet. We use the canonical list of emotions from Paul Ekman:
anger, disgust, fear, happiness, sadness, surprise

<tone>
sarcastic: really any form or irony.
emphatic: intense, highly emotional speech.
vulgar: very familiar speech, potentially some insults

<intensity>
The intensity of the complaint:
low: mild complaint
medium: clear frustration
high: strong language, highly emotional
extreme: genuine distress signals

<intent>
service request, complaint, alert, praise, humor, solidarity, practical info

Please only return the annotations field that are valid for this tweet using xml-like tags.

For instance with a short tweet about a health event on line 1 it would be structured like this:
<event>Someone has a heart attack in the station</event>
<location>Saint-Paul station</location>
<time>early morning</time>
<persona>A middle-aged woman from Provence</persona>
<language_register>casual</language_register>
<language_quality>standard</language_quality>
<emotion>fear</emotion>
<intensity>high</intensity>
<theme>health incident</theme>

Instead with a sex agression, we could have something like this:
<event>The author of the tweet is victim of a sexual agression</event>
<location>Garge de Cergy-Pontoise</location>
<time>November 9th</time>
<persona>A red-headed woman</persona>
<language_register>formal</language_register>
<language_quality>standard</language_quality>
<language_switching>french, english</language_switching>
<emotion>sadness</emotion>
<intensity>high</intensity>
<intent>alert</intent>
<theme>sexual violence</theme>

### Analysis ###
[Short analysis/recontextualization of the tweet that should ultimately converge on the features]

### Annotation ###
[The final annotation using the standard fields without anything else. Especially, DO NOT include the text]
\end{verbatim}
\caption{Prompt for linguistic annotation.}
\label{fig:prompt-2}
\end{figure}

%% file: tex/prompts/draft-generation.tex
\begin{figure}[t]
\scriptsize
\centering
\begin{verbatim}
You have been submitted this tweet:
{text}
  
This tweet has been annotated like this:
{annotation}
  
The annotation follows on a specific set of guidelines.

<event>
A description of the event related in the tweet in English.

<location>
If indicated, name of the location(s) where the event take place.

<time>
If indicated/specified, time of the day or the year the event take place.

<persona>
If specified, a short description of what the author of the tweet could be based on text clues.

<language_register>
Standardized values can be either:
formal: official complaints, institutional tone, requests for attestations
standard: standardized conversational tone in French, neutral tone, proper grammar,
casual: colloquial form of French for online discussions ("mdr", "vzy", contractions like "j'suis", "y'a")
street: argot, verlan like "ouf", "chelou"
dialectical: regional variety of French, not necessarily in France (also include pidgin etc.)

<language_quality>
standard: generally correct French
imperfect: a few typo, still acceptable for conversational style
poor: hard to parse, showing visible struggle with written French

<language_switching>
If valid, simply list the languages that used in lowercase. If French and English: french, english etc.

<emotion>
The primary emotion conveyed by the tweet. We use the canonical list of emotions from Paul Ekman:
anger, disgust, fear, happiness, sadness, surprise

<tone>
sarcastic: really any form or irony.
emphatic: intense, highly emotional speech.
vulgar: very familiar speech, potentially some insults

<intensity>
The intensity of the complaint:
low: mild complaint
medium: clear frustration
high: strong language, highly emotional
extreme: genuine distress signals

<intent>
service request, complaint, alert, praise, humor, solidarity, practical info

Finally you also have a specific entry distress:

<distress>
absent
present

**Criteria for present:**
<...>

**Criteria for absent:**
<...>

Now you have to create a draft leading to the annotation, from the tweets. Even though you already have the final annotations, 
it should really be as if you are just focused on extrapolating them from the tweet.
  
The draft should avoid excessive bold and bullet point, be rather structured like a well stylized text flow. 
It should condensed and to the point, not overdo it.
  
Before starting the draft, you should briefly assess the task at hand, based on the annotations to predict and the text as input. 
You should especially estimate how long the draft should be.
  
You should format your answer like this:
    
<analysis>
[short analysis of the way you would structure the draft and assess its length given the complexity of the tweet.]
</analysis>

<draft>
[your draft leading to the tweet]
</draft>
\end{verbatim}
\caption{Draft generation prompt. Criteria for present/absent are similar to those in Figure~\ref{fig:prompt-1}.}
\label{fig:prompt-3}
\end{figure}